\documentstyle[colap]{article}

\newcounter{examplectr}
\newcounter{subexamplectr}
\newenvironment{exo}%
   {\addtocounter{examplectr}{1}
     \setcounter{subexamplectr}{0}
     \begin{list}
       {(\arabic{examplectr})}%
       {\setlength{\topsep}{0in}
        \setlength{\leftmargin}{.2in}
               \setlength{\labelsep}{0.025in}}
       \item \begin{minipage}[t]{6cm}
   }%
   {\end{minipage}\end{list}}
\newenvironment{ex}%
   {\addtocounter{examplectr}{1}
     \setcounter{subexamplectr}{0}
     \begin{list}
       {(\arabic{examplectr})}%
       {\setlength{\topsep}{0in}
        \setlength{\leftmargin}{.2in}
               \setlength{\labelsep}{0.025in}}
       \item \begin{minipage}[t]{6.9cm}
   }%
   {\end{minipage}
    \end{list}
    \ifnum\thesubexamplectr<1 \else{\vspace*{\parskip}\mbox{}} \fi
   }

\newenvironment{subex}%
   { \addtocounter{subexamplectr}{1}
     \begin{list}
       {\alph{subexamplectr}.}%
       {\setlength{\topsep}{-\parskip}
        \setlength{\leftmargin}{0.15in}
        \setlength{\labelsep}{0.025in}}
       \item
   }%
   {\end{list}}

\newcommand{\exnum}[2]{\addtocounter{examplectr}{#1}(\arabic{examplectr}{#2})\addtocounter{examplectr}{-#1}}

\newcommand{\VM}{\mbox{{\sf Verb}{\sl mobil\/}}}

\newcommand{\eg}{\mbox{e.g.} }

\begin{document}

\author{Michael Dorna \and Martin C. Emele\\[2ex]
  Institut f\"{u}r Maschinelle Sprachverarbeitung\\
  Azenbergstra{\ss}e 12\\
  D-70174 Stuttgart\\
  {\small\verb|\{dorna,emele\}@ims.uni-stuttgart.de|}\\[1ex]
  {\em Published in Proceedings of COLING '96.}}

\date{\today}

\title{Semantic-based Transfer\thanks{This work was funded by the
    German Federal Ministry of Education, Science, Research and
    Technology (BMBF) in the framework of the \VM\ project under grant
    01~IV~101~U. We would like to thank our colleagues of the \VM\ 
    subproject Transfer, our IMS colleagues Ulrich Heid and C.J. Rupp
    and our anonymous reviewers for useful feedback and discussions on
    earlier drafts of the paper. The responsibility for the contents
    of this paper lies with the authors.}}

\maketitle

\begin{abstract}
  This article presents a new semantic-based transfer approach
  developed and applied within the \VM\ Machine Translation project.
  We give an overview of the declarative transfer formalism
  together with its procedural realization. Our approach is discussed
  and compared with several other approaches from the MT literature.
  The results presented in this article have been implemented and
  integrated into the \VM\ system.
\end{abstract}

\section{Introduction}\label{intro}

\noindent The work presented in this article was developed within the
\VM\ project \cite{Kayetal:94,Wahlster:93}. This is one of the largest
projects dealing with Machine Translation (MT) of spoken language.
Approximately 100 researchers in 29 public and industrial institutions
are involved. The application domain is spontaneous spoken language in
face-to-face dialogs. The current scenario is restricted to the task
of appointment scheduling and the languages involved are English,
German and Japanese.

This article describes the realization of a transfer approach based on
the proposals of \cite{AbbBuschbeck:95,CaspariSchmid:94} and
\cite{Copestake:95}. Transfer-based MT\footnote{For a more detailed
  overview of different approaches to MT, see \eg
  \cite{HutchinsSomers:92}.}, see \eg
\cite{VauquoisBoitet:85,Nagaoetal:85}, is based on contrastive
bilingual corpus analyses from which a bilingual lexicon of transfer
equivalences is derived. In contrast to a purely lexicalist approach
which relates bags of lexical signs, as in Shake-and-Bake MT
\cite{Beaven:92,Whitelock:92}, our transfer approach operates on the
level of semantic representations produced by various analysis steps.
The output of transfer is a semantic representation for the target
language which is input to the generator and speech synthesis to
produce the target language utterance. Our transfer equivalences
abstract away from morphological and syntactic idiosyncracies of
source and target languages.  The bilingual equivalences are described
on the basis of semantic representations.

Since the \VM\ domain is related to discourse rather than isolated
sentences the model theoretic semantics is based on Kamp's Discourse
Representation Theory, DRT \cite{KampReyle:93}.  In order to allow for
underspecification, variants of Underspecified Discourse
Representation Structures (UDRS) \cite{Reyle:93} are employed as
semantic formalisms in the different analysis components
\cite{Bosetal:96a,EggLebeth:95b,Copestakeetal:95}.

Together with other kinds of information, such as tense, aspect,
prosody and morpho-syntax, the different semantic representations are
mapped into a single multi-dimensional representation called \VM\ 
Interface Term (VIT) \cite{VIT:96}. This single information structure
serves as input to semantic evaluation and transfer. The transfer
output is also a VIT which is based on the semantics of the English
grammar (cf. \newcite{Copestakeetal:95}) and used for generation (see
\newcite{KilgerFinkler:95} for a description of the generation
component).

Section~\ref{semantics} of this paper sketches the semantic
representations we have used for transfer. In section \ref{trans} we
introduce transfer rules and discuss examples. In
section~\ref{discuss} we compare our approach with other MT
approaches. In section~\ref{impl} we present a summary of the
implementation aspects. For a more detailed discussion of the
implementation of the transfer formalism see \newcite{DornaEmele:96b}.
Finally, section~\ref{sum} summarizes the results.

\section{Semantic Representations}\label{semantics}

The different \VM\ semantic construction components use variants of
UDRS as their semantic formalisms, cf.
\cite{Bosetal:96a,EggLebeth:95b,Copestakeetal:95}. The ability to
underspecify quantifier and operator scope together with certain
lexical ambiguities is important for a practical machine translation
system like \VM\ because it supports ambiguity preserving
translations. The disambiguation of different readings could require
an arbitrary amount of reasoning on real-world knowledge and thus
should be avoided whenever possible.

In the following examples we assume an explicit event-based semantics
\cite{Dowty:89,Parsons:91} with a Neo-Davidsonian representation of
semantic argument relations. All semantic entities in UDRS are
uniquely labeled. A label is a pointer to a semantic predicate making
it easy to refer to. The labeling of all semantic entities allows a
flat representation of the hierarchical structure of argument and
operator and quantifier scope embeddings as a set of labeled
conditions. The recursive embedding is expressed via additional
subordination constraints on labels which occur as arguments of such
operators.

Example \exnum{+1}{a} shows one of the classical \VM\ examples and its
possible English translation \exnum{+1}{b}.

\begin{ex}
\begin{subex}
  {\em Das pa\ss t echt schlecht bei mir.}
\end{subex}
\begin{subex}
  {\em That really doesn't suit me well.}
\end{subex}
\end{ex}

\noindent The corresponding semantic representations are given in
\exnum{+1}{a} and \exnum{+1}{b}, respectively.\footnote{For
  presentation purposes we have simplified the actual VIT
  representations.}

\begin{ex}
\begin{subex}
  {\small
    \verb|[l1:echt(l2), l2:schlecht(i1),|\\ 
    \verb| l3:passen(i1), l3:arg3(i1,i2),|\\ 
    \verb| l4:pron(i2), l5:bei(i1,i3), l6:ich(i3)]|}
\end{subex}
\begin{subex}
  {\small
    \verb|[l1:real(l2), l2:neg(l7), l7:good(i1),|\\ 
    \verb| l3:suit(i1), l3:arg3(i1,i2),|\\
    \verb| l4:pron(i2), l5:arg2(i1,i3), l6:ego(i3)]|}
\end{subex}
\end{ex}

\noindent
Semantic entities in \exnum{0}{} are represented as a Prolog list of
labeled conditions. After the unification-based semantic construction,
the logical variables for labels and markers, such as events, states
and individuals, are skolemized with special constant symbols, \eg
{\tt l1} for a label and {\tt i1} for a state. Every condition is
prefixed with a label serving as a unique identifer. Labels are also
useful for grouping sets of conditions, \eg for partitions which
belong to the restriction of a quantifier or which are part of a
specific sub-DRS.  Additionally, all these special constants can be
seen as pointers for adding or linking information within and between
multiple levels of the VIT.

Only the set of semantic conditions is shown in \exnum{0}{}; the other
levels of the multi-dimensional VIT representation, which contain
additional semantic, pragmatic, morpho-syntactic and prosodic
information, have been left out here. If necessary, such additional
information can be used in transfer and semantic evaluation for
resolving ambiguities or in generation for guiding the realization
choices. Furthermore, it allows transfer to make fine-grained
distinctions between alternatives in cases where the semantic
representations of source and target language do not match up exactly.

Semantic operators like negation, modals or intensifier adverbials,
such as {\em really}, take extra label arguments for referring to
other elements in the flat list which are in the relative scope of
these operators.\footnote{For the concrete example at hand, the
  relative scope has been fully resolved by using the explicit labels
  of other conditions. If the scope were underspecified, explicit
  subordination constraints would be used in a special scope slot of
  the VIT. The exact details of subordination are beyond the scope of
  this paper, cf.~\newcite{FrankReyle:95} and \newcite{Bosetal:96a}
  for implementations.}

This form of semantic representation has the following advantages for
transfer:
\begin{itemize}
\item It is possible to preserve the underspecification of quantifier
  and operator scope if there is no divergence regarding scope
  ambiguity between source and target languages.
\item Coindexation of labels and markers in the source and target
  parts of transfer rules ensures that the semantic entities are
  correctly related and hence obey any semantic constraints which may
  be linked to them.
\item To produce an adequate target utterance additional constraints
  which are important for generation, \eg sortal, topic/focus
  constraints etc., may be preserved.
\item There need not be a $1:1$ relation between semantic entities and
  individual lexical items. Instead, lexical units may be decomposed
  into a set of semantic entities, \eg in the case of derivations and
  for a more fine grained lexical semantics. Lexical decomposition
  allows us to express generalizations and to apply transfer rules to
  parts of the decomposition.
\end{itemize}

\section{Our Transfer Approach}\label{trans}

\noindent
Transfer equivalences are stated as relations between sets of source
language (SL) and sets of target language (TL) semantic entities. They
are usually based on individual lexical items but might also involve
partial phrases for treating idioms and other collocations, \eg
verb-noun collocations (see example \exnum{+6}{} below). After
skolemization of the semantic representation the input to transfer is
variable free. This allows the use of logical variables for labels and
markers in transfer rules to express coindexation constraints between
individual entities such as predicates, operators, quantifiers and
(abstract) thematic roles. Hence the skolemization prevents unwanted
unification of labels and markers while matching individual transfer
rules against the semantic representation.

The general form of a transfer rule is given by
\begin{center}\verb|SLSem, SLConds TauOp TLSem, TLConds.|\end{center}
where {\tt SLSem} and {\tt TLSem} are sets of semantic entities. {\tt
  TauOp} is an operator indicating the intended application direction
(one of {\tt <->,->,<-}). {\tt SLConds} and {\tt TLConds} are optional
sets of SL and TL conditions, respectively. All sets are written as
Prolog lists and optional conditions can be omitted.

On the source language, the main difference between the {\tt SLSem}
and conditions is that the former is matched against the input and
replaced by the {\tt TLSem}, whereas conditions act as filters on the
applicability of individual transfer rules without modifying the input
representation. Hence conditions may be viewed as general inferences
which yield either true or false depending on the context. The context
might either be the local context as defined by the current VIT or the
global context defined via the domain and dialog model. Those
inferences might involve arbitrarily complex inferences like anaphora
resolution or the determination of the current dialog act. In an
interactive system one could even imagine that conditions are posed as
yes/no-questions to the user to act as a negotiator \cite{Kayetal:94}
for choosing the most plausible translation.

If the translation rules in \exnum{+1}{} are applied to the semantic
input in \exnum{0}{a} they yield the semantic output in \exnum{0}{b}.
We restrict the following discussion to the direction from German to
English but the rules can be applied in the other direction as well.

\begin{ex}
\begin{subex}
  {\small\verb|[L:echt(A)] <-> [L:real(A)].|}
\end{subex}
\begin{subex}
  {\small
    \makebox[6.5cm]{\verb|[L:passen(E),L:arg3(E,Y),L1:bei(E,X)]<->|}\\ 
    \verb|[L:suit(E),L:arg2(E,X),L:arg3(E,Y)].|}
\end{subex}
\begin{subex}
  {\small \verb|[L:schlecht(E)],[L1:passen(E)] <->|\\ 
    \verb|[L:neg(A),A:good(E)].|}
\end{subex}
\begin{subex}
  {\small\verb|[L:ich(X)] <-> [L:ego(X)].|}
\end{subex}
\begin{subex}
  {\small\verb|[L:pron(X)] <-> [L:pron(X)].|}
\end{subex}
\end{ex}

\noindent The simple lexical transfer rule in \exnum{0}{a} relates the
German intensifier {\tt echt\/} with the English {\tt
  real}\footnote{The semantic predicate {\tt real} abstracts away from
  the adjective/adverbial distinction.}. The variables {\tt L} and {\tt
  A} ensure that the label and the argument of the German {\tt echt}
are assigned to the English predicate {\tt real}, respectively.

The equivalence in \exnum{0}{b} relates the German predicate {\tt
  passen\/} with the English predicate {\tt suit}. The rule not only
identifies the event marker {\tt E}, but unifies the instances {\tt X}
and {\tt Y} of the relevant thematic roles. Despite the fact that the
German {\em bei\/}-phrase is analysed as an adjunct, it is treated
exactly like the argument {\tt arg3\/} which is syntactically
subcategorized. This rule shows how structural divergences can easily
be handled within this approach.

\begin{exo}
  {\small 
    \verb|[L:passen(E), L1:bei(E,X)] <->|\\ 
    \verb|[L:suit(E),   L:arg2(E,X)].|}
\end{exo}

\noindent The rule in \exnum{-1}{b} might be further abbreviated to
\exnum{0}{} by leaving out the unmodified {\tt arg3}, because it is
handled by a single metarule, which passes on all semantic entities
that are preserved between source and target representation. This also
makes the rule for \exnum{-1}{e} superfluous, since it uses an
interlingua predicate for the anaphor in German and English.

The rule in \exnum{-1}{c} illustrates how an additional condition
({\tt [L1:passen(E)]}) might be used to trigger a specific translation
of {\em schlecht\/} into {\em not good\/} in the context of {\em
  passen\/}. The standard translation of {\em schlecht\/} to {\em
  bad\/} is blocked for verbs like {\em suit}, that presuppose a
positive attitude adverbial.\footnote{Instead of using a specific
  lexical item like {\em passen} the rule should be abstracted for a
  whole class of verbs with similar properties by using a type
  definition, \eg {\tt
    type(de,pos\_attitude\_verbs,[gehen,passen,\ldots])}. For a
  description of type definitions see \exnum{+7}{} below.} One main
advantage of having such conditions is the preservation of the
modularity of transfer equivalences because we do not have to specify
the translation of the particular verb which only triggers the
specific translation of the adverbial. Consequently, the transfer
units remain small and independent of other elements, thus the
interdependencies between different rules are vastly reduced.  The
handling of such rule interactions is known to be one of the major
problems in scaling up MT systems.

A variation on example \exnum{-3}{} is given in \exnum{+1}{}.

\begin{ex}
\begin{subex}
  {\em Das pa\ss t mir echt schlecht.}
\end{subex}
\begin{subex}
  {\em That really doesn't suit me well.}
\end{subex}
\end{ex}

\noindent The translation is exactly the same, but the German verb
{\em passen} takes an indirect object {\em mir} instead of the adjunct
{\em bei}-phrase in \exnum{-4}{}. The appropriate transfer rule looks
like \exnum{+1}{a} which can be reduced to \exnum{+1}{b} because no
argument switching takes place and we can use the metarule again.

\begin{ex}
\begin{subex}
{\small
\makebox[6.5cm]{\verb|[L:passen(E),L:arg2(E,X),L:arg3(E,Y)]<->|}\\
\verb|[L:suit(E),  L:arg2(E,X),L:arg3(E,Y)].|}
\end{subex}
\begin{subex}
{\small\verb|[L:passen(E)] <-> [L:suit(E)].|}
\end{subex}
\end{ex}

\noindent In a purely monotonic system without overriding it would be possible
to apply the transfer rule in \exnum{0}{b} to sentence \exnum{-5}{} in
addition to the rule in \exnum{-2}{} leading to a wrong translation.
Whereas in the underlying rule application scheme assumed here, the
more general rule in \exnum{0}{b} will be blocked by the more specific
rule in \exnum{-2}{}.

The specificity ordering of transfer rules is primarily defined in
terms of the cardinality of matching subsets and by the subsumption
order on terms.  In addition, it also depends on the cardinality and
complexity of conditions.  For the {\em passen} example at hand, the
number of matching predicates in the two competing transfer rules
defines the degree of specificity.

The following example illustrates how conditions are used to enforce
selectional restrictions from the domain model. For example {\it
  Termin\/} in German might either be translated as {\it
  appointment\/} or as {\it date}, depending on the context.

\begin{ex}
\begin{subex}
{\small\verb|[L:termin(X)] <-> [L:appointment(X)].|}
\end{subex}
\begin{subex}
{\small
\verb|[L:termin(X)],|\\
\verb|[sort(X)=<~temp_point] <-> [L:date(X)].|}
\end{subex}
\end{ex}

\noindent
The second rule \exnum{0}{b} is more specific, because it uses an
additional condition. This rule will be tried first by calling the
external domain model for testing whether the sort assigned to {\tt X}
is not subsumed by the sort \verb|temp_point|. Here, the first rule
\exnum{0}{a} serves as a kind of default with respect to the
translation of {\it Termin\/}, in cases where no specific sort
information on the marker {\tt X} is available or the condition in rule
\exnum{0}{b} fails.

In \exnum{+1}{}, a light verb construction like {\em einen
  Terminvorschlag machen\/} is translated into {\em suggest a date} by
decomposing the compound and light verb to a simplex verb and its
modifying noun.

\begin{exo}
{\small
  \verb|[L:machen(E),L:arg3(E,X),|\\
  \verb| L1:terminvorschlag(X)] <->|\\
  \verb|[L:suggest(E),L:arg3(E,X),L1:date(X)].|}
\end{exo}

\noindent
We close this section with a support verb example \exnum{+1}{} showing
the treatment of head switching in our approach. The German
comparative construction {\em lieber sein\/} (lit.:{\em be more
  liked}) in \exnum{+1}{a} is translated by the verb {\em prefer} in
\exnum{+1}{b}.

\begin{ex}
  \begin{subex}
    {\em Dienstag ist mir lieber.}
  \end{subex}
  \begin{subex}
    {\em I would prefer Tuesday.}
  \end{subex}
\end{ex}

\begin{exo}
{\small
  \verb|[L:support(S,L1),L2:experiencer(S,X)|\\
  \verb| L1:lieb(Y),L1:comparative(Y)] <->|\\
  \verb|[L:prefer(S),L:arg1(S,X),L:arg3(S,Y)].|}
\end{exo}

\noindent
The transfer rule in \exnum{0}{} matches the decomposition of the
comparative form {\em lieber} into its positive form {\em lieb} and an
additional comparative predicate together with the support verb {\em
  sein} such that the comparative construction {\em lieber sein} ({\tt
  Y} {\em ist\/} {\tt X} {\em lieber}) is translated as a whole to the
English verb {\em prefer} ({\tt X} {\em prefers} {\tt Y}).

\section{Discussion}\label{discuss}

The main motivation for using a semantic-based approach for transfer
is the ability to abstract away from morphological and syntactic
idiosyncrasies of individual languages. Many of the traditional cases
of divergences discussed, \eg by \newcite{Dorr:94}, are already
handled in the \VM\ syntax-semantics interface, hence they do not show
up in our transfer approach. Examples include cases of categorial and
thematic divergences. These are treated in the linking between
syntactic arguments and their corresponding thematic roles.

Another advantage of a semantic-based transfer approach over a pure
interlingua approach, \eg \newcite{Dorr:93}, or a direct structural
correspondence approach, \eg \newcite{Slocumetal:87}, is the gain in
modularity by allowing language independent grammar development.
Translation equivalences relating semantic entities of the source and
target grammars can be formulated in a grammar independent bilingual
semantic lexicon. In cases where the semantic representations of
source and target language are not isomorphic, a nontrivial transfer
relation between the two representations is needed. But it is clearly
much easier to map between flat semantic representations than between
either syntactic trees or deeply nested semantic representations

An interlingua approach presumes that a single representation for
arbitrary languages exists or can be developed. We believe from a
grammar engineering point of view it is unrealistic to come up with
such an interlingua representation without a strict coordination
between the monolingual grammars. In general, a pure interlingua
approach results in very application and domain specific knowledge
sources which are difficult to maintain and extend to new languages
and domains. This holds especially in the \VM\ context with its
distributed grammar development.

Whereas our approach does not preclude the use of interlingua
predicates. We use interlingua representations for time and date
expressions in the \VM\ domain. Similarly for prepositions, cf.
\newcite{BuschbeckNuebel:95}, it makes sense to use more abstract
relations which express fundamental relationships like {\em temporal
  location\/} or {\em spatial location}. Then it is left to the
language specific grammars to make the right lexical choices.  

\begin{ex}
\begin{subex}
\verb|type(de,temp_loc,[an,in,um,zu]).|
\end{subex}
\begin{subex}
{\small{\em am} Dienstag, {\em im} Mai, {\em um} drei, {\em zu} Ostern}
\end{subex}
\begin{subex}
\verb|type(en,temp_loc,[on,in,at]).|
\end{subex}
\begin{subex}
{\small{\em on} Tuesday, {\em in} May, {\em at} three, {\em at} Easter}
\end{subex}
\end{ex}

\noindent The class definitions in \exnum{0}{a} and \exnum{0}{c}
cluster together those prepositions which can be used to express a
temporal location.  The names \verb|de| and \verb|en| are the SL and
TL modules in which the class is defined, \verb|temp_loc| is the class
name and the list denotes the extension of the class. \exnum{0}{b} and
\exnum{0}{d} show possible German and English lexicalizations.

\begin{exo}
  \verb|[temp_loc(E,X)],[sort(X)=<time] <->|\\
  \verb|[temp_loc(E,X)].|
\end{exo}

\noindent
The interlingua rule in \exnum{0}{} identifies the abstract temporal
location predicates under the condition that the internal argument is
more specific than the sort {\tt time}.  This condition is necessary
because of the polysemy of those prepositions.  During compilation the
SL class definition will be automatically expanded to the individual
predicates, whereas the TL class definition will be kept unexpanded
such that the target grammar might be able to choose one of the
idiosyncratic prepositions.

\noindent Mixed approaches like \newcite{Kaplanetal:89} can be characterized by
mapping syntax as well as a predicate-argument structure
(f-structure).  As already pointed out, \eg in
\cite{SadlerThompson:91}, this kind of transfer has problems with its
own multiple level mappings, \eg handling of verb-adverb head
switching, and does not cleanly separate monolingual from contrastive
knowledge, either.  In \newcite{KaplanWedekind:93} an improved
treatment of head switching is presented but it still remains a less
general solution.

A semantic approach is much more independent of different syntactic
analyses which are the source of a lot of classical translation
problems such as structural and categorial divergences and mismatches.
In our approach grammars can be developed for each language
independently of the transfer task and can therefore be reused in
other applications.

At first glance, our approach is very similar to the semantic transfer
approach presented in \newcite{Alshawietal:91}.  It uses a level of
underspecified semantic representations as input and output of
transfer. The main differences between our approach and theirs are the
use of flat semantic representations and the non-recursive transfer
rules. The set-oriented representation allows much simpler operations
in transfer for accessing individual entities (set membership) and for
combining the result of individual rules (set union). Furthermore,
because the recursive rule application is not part of the rules
themselves, our approach solves problems with discontinuous
translation equivalences which the former approach cannot handle well.
A transfer rule for such a case is given in \exnum{-8}{}.

Our current approach is strongly related to the Shake-and-Bake
approach of \newcite{Beaven:92} and \newcite{Whitelock:92}. But
instead of using sets of lexical signs, {i.e.} morpho-syntactic {\em
  lexemes} as in Shake-and-Bake, we specify translation equivalences
on sets of arbitrary {\em semantic entities}.  Therefore, before
entering the transfer component of our system, individual lexemes can
already be decomposed into sets of such entities, \eg for stating
generalizations on the lexical semantics level or providing suitable
representations for inferences.  For example, the wh-question word
{\em when} is decomposed into \verb|temp_loc(E,X), whq(X,R),
time(R,X)| (lit.: {\em at which time}), hence no additional transfer
rules are required.  Similarly, German composita like {\em
  Terminvorschlag} are decomposed into its compounds, \eg
\verb|termin(i2), n_n(i1,i2), vorschlag(i1)| where \verb|n_n| denotes
a generic noun-noun relation.  As a result a compositional translation
as {\em proposal for a date} is possible without stating any
additional translation equivalences to the ones for the simplex nouns.

Another major difference is the addition of conditions which trigger
and block the applicability of individual transfer rules. For instance
in the specific translation of {\em schlecht} to {\em not good} as
defined in \exnum{-9}{c}, without conditions, one would have to add
the verb {\em passen} into the bag to test for such a specific
context.  As a consequence the translation of the verb needs to be
reduplicated, whereas in our approach, the translation of the verb can
be kept totally independent of this specific translation of the
adverbial, because the condition functions merely as a test.  

These examples also illustrates the usefulness of labeled conditions,
because the negation operator can take such a label as an argument and
we can use unification again to achieve the correct coindexation.  If
we would use a hierarchical semantics instead, as in the original
Shake-and-Bake aproach, where the negation operator embeds the verb
semantics we would have to translate {\tt schlecht(e), passen(e)\/}
into {\tt not(suit(e), well(e))} in one rule because there is no
coindexation possible to express the correct embedding without the
unique labeling of predicates.

Finally, we have filled the lack of an adequate control strategy for
Shake-and-Bake by developing a nonmonotonic control strategy which
orders more specific rules before less specific ones.  This strategy
allows the specification of powerful default translations. Whereas
without such an ordering special care is needed to prevent a
compositional translation in cases where a more specific
noncompositional translation also exists.

The same argument about control holds in comparison to the
unification-based transfer approach on Mimimal Recursion Semantics
(MRS) \cite{Copestakeetal:95,Copestake:95}. In addition, we use
matching on first order terms instead of feature structure
unification. Full unification might be problematic because it is
possible to add arbitrary information during rule application, \eg by
further unifying different arguments.  The other main difference is
our nonmonotonic control component whereas the MRS approach assumes a
monotonic computation of all possible transfer equivalences which are
then filtered by the generation grammar. It is difficult to judge the
feasibility of their approach given the fact that only a limited
coverage has been addressed so far.

\section{Implementation}\label{impl}

A more detailed presentation of the implementation aspects of our
transfer approach can be found in \newcite{DornaEmele:96b}. The
current transfer implementation consists of a transfer rule compiler
which takes a set of rules like the one presented in
section~\ref{trans} and compiles them into two executable Prolog
programs one for each translation direction. The compiled program
includes the selection of rules, the control of rule applications and
calls to external processes if necessary.

Because both the transfer input and the matching part of the rules
consist of sets we can exploit ordered set operations during
compilation as well as at runtime to speed up the matching process and
for computing common prefixes which are shared between different
rules.

The compiled transfer program is embedded in the incremental and
parallel architecture of the \VM\ Prototype. Interaction with external
modules, \eg the domain model and dialog module or other inference
components, is done via a set of predefined abstract interface
functions which may be called in the condition part of transfer rules.
The result is a fully transparent and modular interface for filtering
the applicability of transfer rules.

\section{Summary}\label{sum}

This paper presents a new declarative transfer rule formalism, which
provides an implementation platform for a semantic-based transfer
approach. This approach combines ideas from a number of recent MT
proposals and tries to avoid many of the well known problems of other
transfer and interlingua approaches.

The declarative transfer correspondences are compiled into an
executable Prolog program. The compiler exploits indexing for more
efficient search of matching rules. There is a nonmonotonic but
rule-independent control strategy based on rule specificity. 

Currently, the transfer component contains about 1700 transfer rules.
Thanks to the set orientation and indexing techniques we did not
encounter any scaling problems and the average runtime performance for
a 15 word sentence is about 30 milliseconds.

Future work will include the automatic acquisition of transfer rules
from tagged bilingual corpora to extend the coverage and an
integration of domain specific dictionaries.

\bibliographystyle{acl}
\begin{scriptsize}

\end{scriptsize}

\end{document}